\documentclass{article}

\usepackage{arxiv}
\usepackage{amsmath}
\usepackage{subcaption}
\usepackage[utf8]{inputenc} 
\usepackage[T1]{fontenc}    
\usepackage{hyperref}       
\usepackage{url}            
\usepackage{booktabs}       
\usepackage{amsfonts}       
\usepackage{nicefrac}       
\usepackage{microtype}      
\usepackage{cleveref}       
\usepackage{lipsum}         
\usepackage{graphicx}
\usepackage{natbib}
\usepackage{doi}
\usepackage{enumitem}

\title{\textsf{Auto}mated \textsf{M}otion \textsf{A}rtifact \textsf{C}heck for \textsf{MRI} (\textit{AutoMAC-MRI}): An Interpretable Framework for Motion Artifact Detection and Severity Assessment}

\newif\ifuniqueAffiliation
\uniqueAffiliationtrue

\ifuniqueAffiliation 
\author{%
	\textbf{Antony Jerald}\\
	\textit{GE HealthCare, Bangalore, INDIA}\\
	\texttt{antony.jerald@gehealthcare.com}\\
	\And
	\textbf{Dattesh Shanbhag}\\
	\textit{GE HealthCare, Bangalore, INDIA}\\
	\texttt{dattesh.shanbhag@gehealthcare.com}\\
	\And
	\textbf{Sudhanya Chatterjee}\\
	\textit{GE HealthCare, Bangalore, INDIA}\\
	\texttt{sudhanya.chatterjee@gehealthcare.com}\\
}
\fi

\renewcommand{\shorttitle}{AutoMAC-MRI}

\begin{document}
\maketitle

\begin{abstract}
Motion artifacts degrade MRI image quality and increase patient recalls. Existing automated quality assessment methods are largely limited to binary decisions and provide little interpretability. We introduce AutoMAC‑MRI, an explainable framework for grading motion artifacts across heterogeneous MR contrasts and orientations. The approach uses supervised contrastive learning to learn a discriminative representation of motion severity. Within this feature space, we compute grade-specific affinity scores that quantify an image’s proximity to each motion grade, thereby making grade assignments transparent and interpretable.
We evaluate AutoMAC‑MRI on more than 5,000 expert‑annotated brain MRI slices spanning multiple contrasts and views. Experiments assessing affinity scores against expert labels show that the scores align well with expert judgment, supporting their use as an interpretable measure of motion severity. By coupling accurate grade detection with per‑grade affinity scoring, AutoMAC‑MRI enables inline MRI quality control, with the potential to reduce unnecessary rescans and improve workflow efficiency.
\end{abstract}


\section{Introduction}

Magnetic resonance imaging (MRI) is a non-invasive modality that delivers excellent soft tissue contrast, enabling detailed structural and functional characterization across neurological, abdominal, and musculoskeletal applications without ionizing radiation. However, the extended acquisition times of many sequences render MRI uniquely susceptible to patient motion—ranging from voluntary head movements and eye blinks to respiration and vascular pulsation—thereby producing characteristic artifacts (e.g., blurring, ringing, and phase encoding ghosting) that can propagate into subsequent analytical stages \cite{zaitsev2015motion,havsteen2017movement}. Even subtle motion has been shown to bias quantitative measurements, inflating variance and distorting morphometric estimates (e.g., cortical thickness and volumetrics) as well as functional and structural connectivity metrics, with documented effects across multiple MRI pipelines \cite{power2012spurious,reuter2015head,alexander2016subtle}. Clinically, motion degraded images frequently necessitate repeat acquisitions, increase direct and indirect costs, and delay reporting and patient throughput, with the impact particularly pronounced in high volume or resource constrained settings \cite{havsteen2017movement}. Accordingly, efficient and interpretable quality control, integrated inline during acquisition and informed by automated image quality metrics, can detect motion early and support decisions by technologists. It also enables immediate reacquisition while the subject remains in the scanner, thereby reducing avoidable rescans and workflow inefficiencies and curbing bias in downstream processing and analysis \cite{esteban2017mriqc,obuchowicz2020magnetic}.

Recent advances in deep learning for medical imaging have spurred a range of CNN methods that automatically detect motion in MR images frequently achieving high accuracy on curated test sets \cite{esses2018automated,vakli2023automatic, jimeno2024automated,eckerintegrating}.  In abdominal imaging, \cite{esses2018automated} trained a CNN to classify T$_2$-weighted liver acquisitions as diagnostic or nondiagnostic, reporting a high negative predictive value for screening nondiagnostic exams. Notably, the evaluation was restricted to a single contrast (T$_2$-weighted liver MRI) and the study did not assess generalization across other sequences.  In T$_1$-weighted brain MRI, \cite{vakli2023automatic} compared end-to-end deep learning with conventional machine learning built on image quality metrics and found similar effectiveness in distinguishing clinically usable vs. unusable scans, also within a binary framework. Importantly, their analysis focused entirely on motion-related artifacts in T$_1$-weighted data and explicitly acknowledges the need for validation on larger and more diverse MRI sequences. Beyond binary classification, another work by \cite{jimeno2024automated} has explored three-class motion detection in T$_1$-weighted brain MRI with lightweight 2D CNNs and Grad-CAM for interpretability. While this approach demonstrate promise, they rely heavily on synthetically motion-corrupted training data and report performance drops on out-of-distribution datasets. Moreover, in this work agreement on prospective data is sensitive to preprocessing choices (e.g., field of view cropping/centering, defacing, co registration) and to the background–anatomy proportion, highlighting the difficulty of achieving robust generalization beyond the development setting.

More recent inline quality control approaches introduced by \cite{eckerintegrating} combine self-supervised contrastive representations for global image quality assessment with Swin U-Net for local artifact detection. However, self-supervised learning is not optimized to enforce separability between clinically meaningful motion severity categories, particularly when artifacts are spatially heterogeneous and the boundary between subtle and severe motion must be maintained. In addition, this work infers quality by comparing a test embedding against fixed exemplar embeddings—high-quality and low-quality references—via cosine similarity. Decisions anchored to contrast or view-specific exemplars can be brittle under protocol contrast variation and broader distribution shifts. These factors motivate a method that couples clinically meaningful motion grading with interpretable severity scoring, enabling robust, contrast-invariant quality control.

Building on these considerations, we present AutoMAC-MRI, a slice-wise, contrast-agnostic framework that grades motion artifacts into three clinically actionable categories (No motion, Subtle motion, Severe motion) and outputs per-slice Motion Grade Affinity Scores. The approach trains a ResNet-18 encoder with supervised contrastive learning to produce embeddings with strong intra-class compactness and inter-class separation and a lightweight MLP head operating on frozen embeddings providing calibrated grade assignments. To make grade estimation interpretable, we utilize the learned feature space to obtain per-grade affinity scores aligned with expert judgment. We evaluate AutoMAC-MRI on brain MR slices spanning multiple contrasts (T1-w, T2-w, PD-w, FLAIR) and orientations (axial, coronal, sagittal, oblique) and benchmark against self-supervised and fully supervised baselines. The combination of accurate grading, interpretable severity scoring supports inline quality control decisions during acquisition, helping reduce avoidable rescans and streamline routine workflow.

\section{Methods}

\subsection{Training Framework}
There are two primary objectives for this work:
\begin{itemize}
	\item Learn to categorize motion in MR images by its severity.
	\item In addition to detecting the motion artifact category, provide an affinity score for the detected motion category.
\end{itemize}
The ability to separate classes well in deep feature space provides reliability. This enables robust classification for categories as well~\cite{chen2020simple,khosla2020supervised}. In this work, we ensure that motion categories are well separated in deep feature space. This representation is then utilized to: (a) perform robust motion category detection, and (b) compute affinity scores for each motion category for a given MR image. 

\subsection*{Motion Grading of MR image and Annotation}
We introduce below the motion categories annotated for the paper. All MR slices were annotated by an experienced MRI application specialist with over 10 years of domain expertise. Annotations were performed at the slice level, as the proposed approach evaluates each 2D slices individually. Each slice was assigned to one of three motion grades:
\begin{itemize}
	\item \textbf{No Motion}: No visible motion artifacts.
	\item \textbf{Subtle Motion}: Minor motion artifacts not severe enough to warrant re-scanning.
	\item \textbf{Severe Motion}: High motion artifacts requiring image re-acquisition.
\end{itemize}
This annotation strategy was designed to align motion severity grading with clinical decision-making, ensuring that the labels reflect actionable consequences rather than purely visual assessment. Some sample image annotations are shown in Figure \ref{fig:sample_annotations} in Appendix.

\subsection*{Training}
\begin{figure}[htb]
	\centering
	\caption{The figure shows the overview of the proposed training and scoring pipeline. 
		(a) The training process consists of two stages: Stage~1 employs supervised contrastive learning to construct a well-separated feature space for motion grades, while Stage~2 trains an MLP classifier on frozen embeddings to predict the motion grade of MR images. 
		(b) For severity scoring, grade-specific template vectors are derived from training data. During inference, the 512-dimensional embedding from Stage~1 is compared against these templates using cosine similarity to compute interpretable motion severity scores.}
	\label{fig:arc}
	
	\begin{subfigure}[c]{0.54\textwidth}
		\centering
		\includegraphics[width=\linewidth]{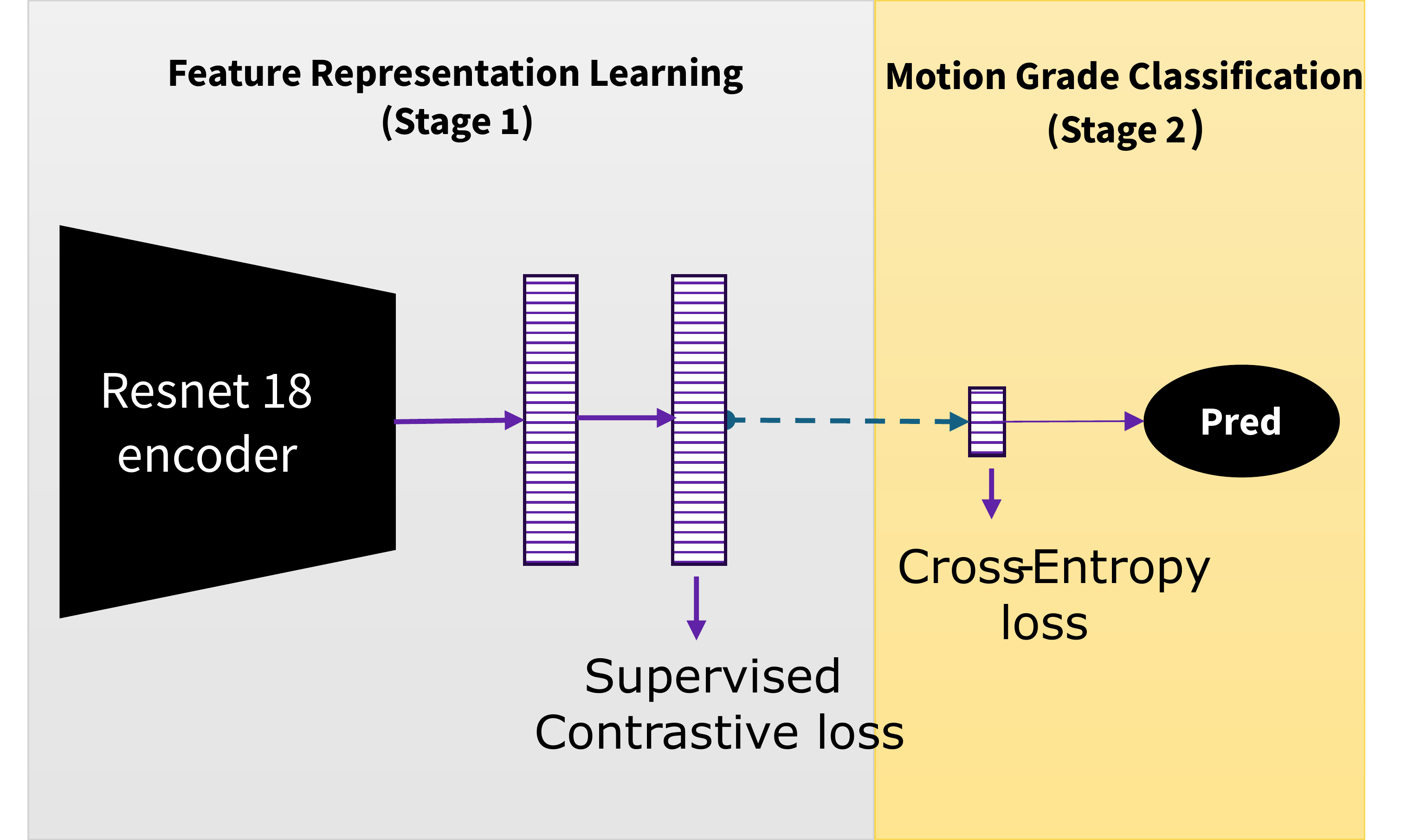}
		\caption{}%
		\label{fig:arc1}
	\end{subfigure}
	\hfill
	\begin{subfigure}[c]{0.45\textwidth}
		\centering
		\includegraphics[width=\linewidth]{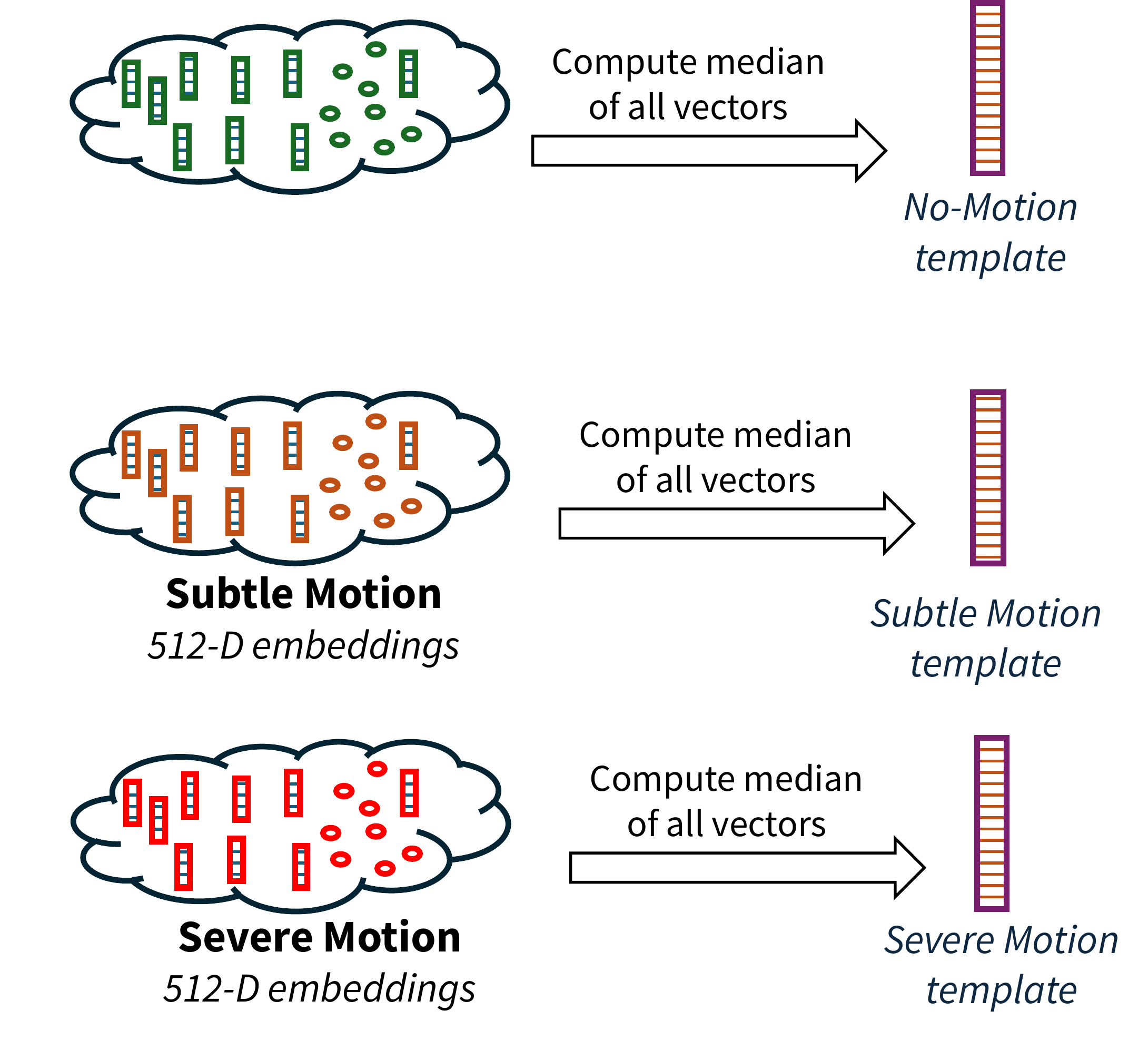}
		\caption{}%
		\label{fig:arc2}
	\end{subfigure}
\end{figure}

Broadly, the proposed method has two training stages. In the first stage we learn separation of the motion categories in deep feature space by training a CNN-based encoder. In the next stage, the CNN-based encoder trained in earlier phase is frozen and a light weight multi-layer perceptron (MLP) is trained to perform a binary motion categorization. Method has been illustrated in Figure~\ref{fig:arc1}.  
\begin{enumerate}[label=\textbf{Stage-\arabic*}, align=left, leftmargin=*]
	
	\item \textit{Feature Representation Learning using Supervised Contrastive Learning}:  
	A ResNet-18 network~\cite{he2016deep}, initialized with ImageNet pre-trained weights, 
	was extended with two additional fully connected layers (each comprising 512 neurons) 
	to form the base encoder network. This augmented network was trained using supervised 
	contrastive learning~\cite{khosla2020supervised} to construct a discriminative feature 
	space for three motion severity categories: \emph{No Motion}, \emph{Subtle Motion}, 
	and \emph{Severe Motion}. The resulting 512-dimensional embeddings were optimized with 
	supervised contrastive loss to maximize intra-class similarity while promoting 
	inter-class separation among the motion grades. This encoder generates a 
	512-length feature vector representation of each image.
	
	\item \textit{Motion Grade Classification using MLP Head}:  
	In the second stage, a lightweight multilayer perceptron (MLP) classifier with 3 neurons 
	was trained on the embeddings generated by the Stage~1 encoder. During this phase, the 
	encoder weights were frozen to preserve the learned feature space, and the classifier 
	was optimized using cross-entropy loss to predict discrete motion grades.
	
\end{enumerate}

\subsection{Motion Grade Affinity Scoring $\left(\text{MoGrAS}\right)$}
As discussed earlier, the classifier head from Stage~2 predicts discrete motion grades using the 512-D embeddings generated by the encoder trained using a supervised contrastive (SupCon) learning regime~\cite{khosla2020supervised}.
The other important objective than predicting the motion grade, is to estimate affinity score of the image for each motion grade. This score adds interpretability to the predicted motion grade. This score is referred to as \textbf{Mo}tion \textbf{Gr}ade \textbf{A}ffinity \textbf{S}core (MoGrAS) hereafter.

The 512-D embeddings learnt from the data using SupCon is used as the basis to compute MoGrAS. The 512-D feature vector from the encoder’s penultimate layer is used for this purpose.

The primary idea is to create feature vector (512-D) templates corresponding to the three motion categories. These templates then act as reference for test data. The 512-D feature vector for a test data is compared against template of each motion grades to obtain respective MoGrAS.

To generate reference templates for each motion grade, we compute the median feature vector for each motion grade across all corresponding training samples (illustrated in Fig.~\ref{fig:arc2}). These template vectors  represent the central tendency of the features in 512-D embedding space for its respective motion grades.
\subsubsection*{Computing MoGrAS during inference}
During inference the 512-D embedding (say $F_{\text{test}}$) for the MR image is obtained using the encoder trained using SupCon as discussed earlier. The affinity score for each motion grade is then obtained by computing cosine similarity between $F_{\text{test}}$ and 512-D embeddings for the motion grade, say $F_\text{g}$, where $\text{g}=\left\{\text{No Motion, Subtle Motion, Severe Motion}\right\}$. Since we have three motion grades, each test MR image shall have three MoGrAS (one corresponding to each motion grade). Hence, for a given grade $g$, the MoGrAS is computed as:

\begin{equation}
	\label{eq:cosine_similarity}
	\text{MoGrAS}_{\text{g}} = 
	\frac{F_{\text{g}} \cdot F_{\text{test}}}{\|F_{\text{g}}\| \, \|F_{\text{test}}\|}
\end{equation}

By virtue of the cosine similarity formulation, MoGrAS ranges from $-1$ to $+1$ (indicating lowest and highest affinity for a motion grade respectively).

\subsection{Dataset Description}
The dataset consisted of 5,304 expert-annotated 2D brain MR slices acquired across multiple imaging contrasts—T1-weighted (T1-w), T2-weighted (T2-w), Proton Density-weighted (PD-w), and FLAIR—and orientations including axial, coronal, sagittal, and oblique. The orientation-wise distribution was 3,183 axial, 1,804 coronal, 313 sagittal, and 4 oblique slices, while the contrast-wise distribution included PD-w (1,135 slices), T1-w (1,879 slices), T2-w (646 slices), and FLAIR (1,645 slices).

To ensure robust model development and evaluation, the dataset was split into training (2,552 slices), validation (478 slices), and testing (2,274 slices) subsets. This stratification was performed while preserving diversity across contrasts and orientations, thus avoiding data bias while training or reporting test statistics.

\subsection{Experiments}
In this section we shall discuss the experiments designed to validate claims in the method section, evaluation of the proposed method with similar approaches and test data performance assessment approach adopted in this work.
\subsubsection{Comparison with other methods}
The proposed method is compared against two other approaches. The methods and basis of their choices are discussed below:

\textbf{Using SimCLR for self-supervised contrastive learning}:  
The self-supervised contrastive learning framework, also known as SimCLR~\cite{chen2020simple} is a popular contrastive loss based self-supervised learning approach. Unlike SimCLR, which uses instance-level contrastive learning without labels, SupCon~\cite{khosla2020supervised} leverages supervised information to pull samples of the same class closer together.

This comparison experiment employs the same encoder architecture (ResNet 18 + MLP) while learning using SimCLR~\cite{chen2020simple} during Stage-1 (hence no label information is used). The Stage-2 of the method was maintained as-it-is.

The primary objective of this experiment is to understand the effectiveness of using SupCon as a self-supervised contrastive learning approach against other similar approaches as SimCLR.

\textbf{Motion grade (3-class) classifier network}:
We compare the proposed method with a standard 3-class classifier trained using cross entropy loss. The exact same architecture is retained for this method as discussed in the Training section (Training Stage 1 + Training Stage 2). In contrast to the proposed method, the entire DL network (encoder and MLP head) are trained in a supervised manner using cross entropy as loss function.

The objective of this experiment is to evaluate whether Stage-1 training in the proposed method (encoder trained using SupCon approach) is more effective at predicting the motion grade.

These comparison methods provide a controlled comparison across distinct learning paradigms, isolating the effect of supervision and contrastive learning on both accuracy and interpretability.

\subsubsection{Feature Space Visualization}
While predicting the motion grade determines the accuracy of the proposed method, it is also important to have a well separated feature space among the motion grades. Such representations ensure robust motion grade detection, and at the same time enable motion grade affinity scoring as discussed in earlier sections.
To assess this, we project the 512-D encoder embeddings (refer Figure~\ref{fig:arc1}) into two dimensions using t-SNE~\cite{maaten2008visualizing}. 

This visualization enables qualitative assessment of cluster formation and separation across motion grades. Well-formed clusters with minimal overlap shall indicate strong intra-class compactness and inter-class discrimination. Such properties are essential for reliable motion grading, especially where boundaries between motion grades are subtle. Comparing visualizations across different training strategies highlights the influence of the learning paradigms on feature space representations.

\subsubsection{Motion Grade Classification and MoGrAS analysis}
The proposed framework provides both discrete motion grading and the affinity score for each motion grade (MoGrAS).

\textit{Interpreting \textnormal{MoGrAS}}: A higher MoGrAS for a particular motion grade indicates a higher alignment of the MR image with the corresponding motion grade.
This continuous scoring mechanism attached to each motion grade complements discrete classification report, enabling nuanced quality assessment and supporting clinically meaningful decision-making.

\section{Results}

\subsection{Feature Space Visualization}
\begin{figure}[htb]
	\centering
	\caption{t-SNE visualization of 512-dimensional embeddings under different training strategies:
		(a) encoder trained using SimCLR,
		(b) fully supervised 3-class DL network, and
		(c) proposed method using supervised contrastive learning.}
	\label{fig:tsne}
	
	\begin{subfigure}{0.33\textwidth}
		\centering
		\includegraphics[width=\linewidth]{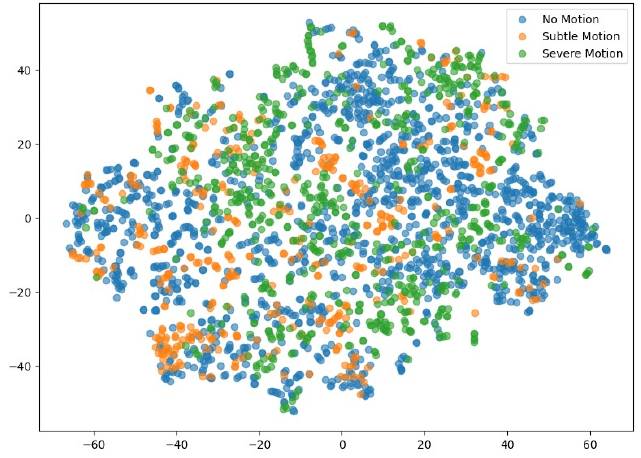}
		\caption{}%
		\label{fig:tsne1}
	\end{subfigure}
	\hfill
	\begin{subfigure}{0.33\textwidth}
		\centering
		\includegraphics[width=\linewidth]{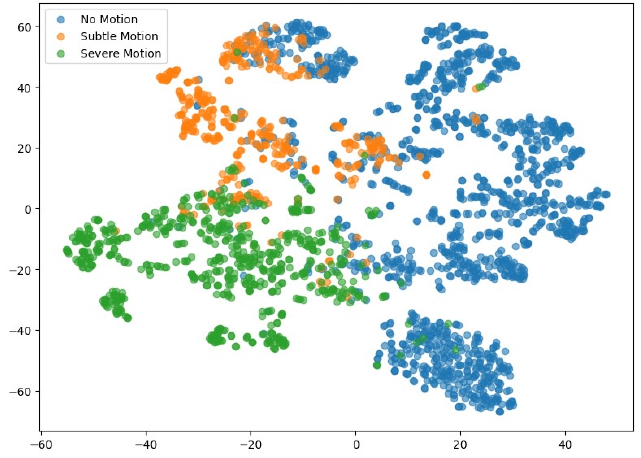}
		\caption{}%
		\label{fig:tsne2}
	\end{subfigure}
	\hfill
	\begin{subfigure}{0.33\textwidth}
		\centering
		\includegraphics[width=\linewidth]{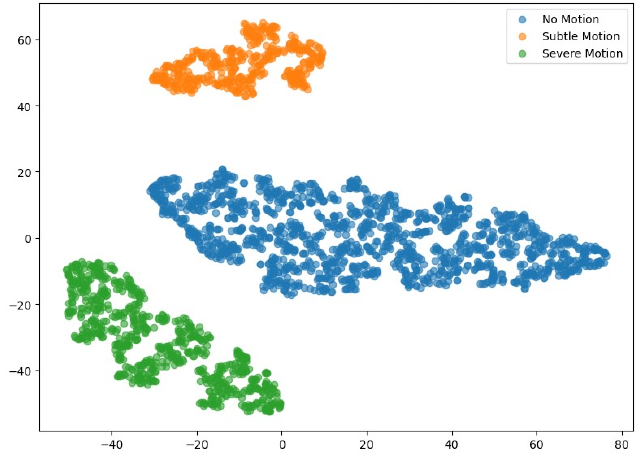}
		\caption{}%
		\label{fig:tsne3}
	\end{subfigure}
\end{figure}

Figure~\ref{fig:tsne} presents t-SNE projections of 512-D embeddings for three motion grades: No Motion, Subtle Motion, and Severe Motion. Results are shown for the three experiments discussed so far: encoder trained using SimCLR (refer Figure~\ref{fig:tsne1}), a fully supervised 3-class DL network (refer Figure~\ref{fig:tsne2}), proposed method (refer Figure~\ref{fig:tsne3}).
The feature space while using SimCLR for Stage-1 training, there is substantial inter-class mixing of features. The fully supervised 3-class classification approach shows reasonable separation, though considerable overlap remains. However for the proposed method (using SupCon for Stage-1 training), across the motion grades we observe distinct clusters with clear boundaries and minimal overlap.

\subsection{Motion Grade Detection Performance}
\begin{figure}[htb]
	\centering
	\caption{Confusion matrices illustrating motion grade classification performance across three training strategies:
		(a) encoder trained using SimCLR,
		(b) fully supervised 3-class DL network, and
		(c) proposed method using supervised contrastive learning.
		Each matrix shows predictions for three motion grades: \emph{No Motion}, \emph{Subtle Motion}, and \emph{Severe Motion}.}
	\label{fig:confusion_matrix}
	
	\begin{subfigure}{0.33\textwidth}
		\centering
		\includegraphics[width=\linewidth]{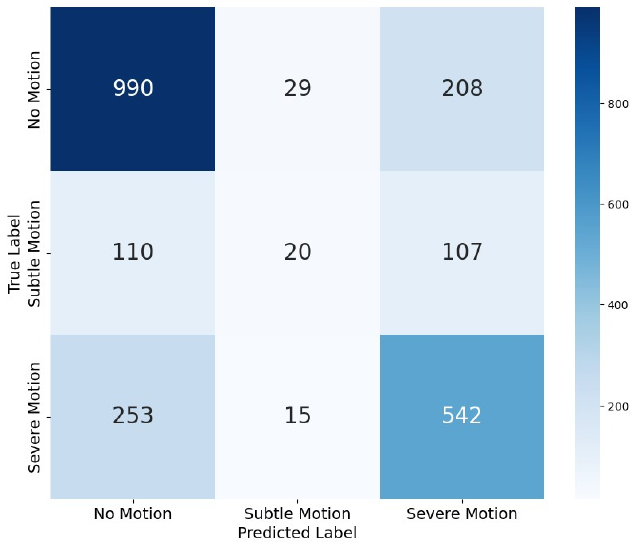}
		\caption{}%
		\label{fig:CM1}
	\end{subfigure}
	\hfill
	\begin{subfigure}{0.33\textwidth}
		\centering
		\includegraphics[width=\linewidth]{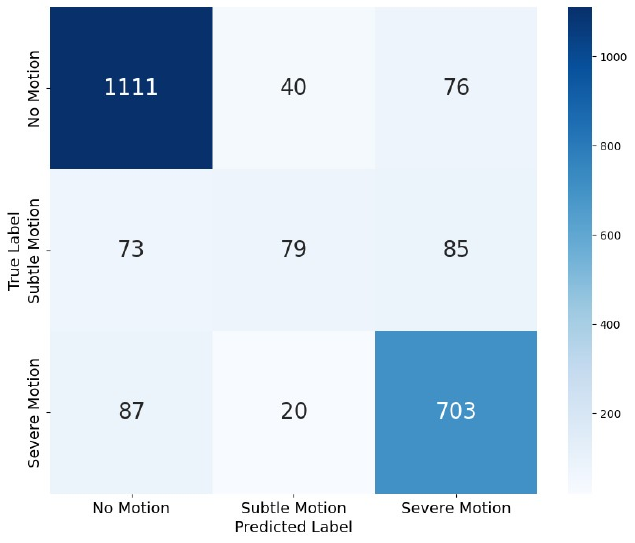}
		\caption{}%
		\label{fig:CM2}
	\end{subfigure}
	\hfill
	\begin{subfigure}{0.33\textwidth}
		\centering
		\includegraphics[width=\linewidth]{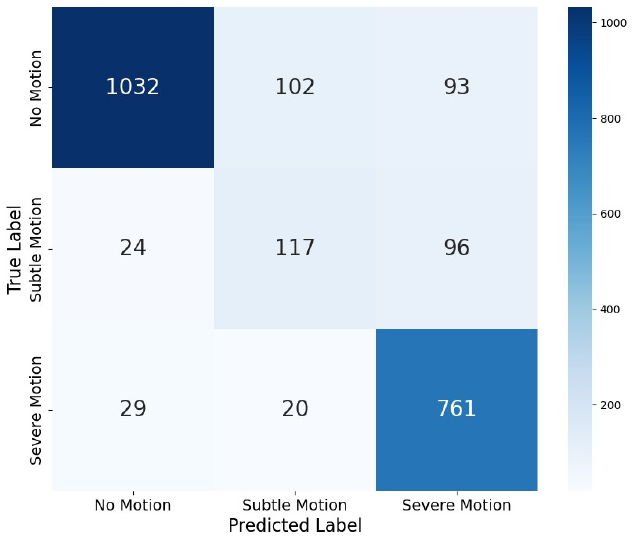}
		\caption{}%
		\label{fig:CM3}
	\end{subfigure}
\end{figure}

The proposed framework was evaluated on a test set comprising 2,274 MR slices (across MRI contrasts and views). Figure~\ref{fig:confusion_matrix} shows confusion matrices for the three training strategies. The accuracy summary is shared in Table~\ref{tab:metrics}. The proposed method achieved the highest overall accuracy \textbf{84.0\%}, compared to \textbf{83.2\%} for the fully supervised 3-class classifier and \textbf{68.2\%} for the experiment where SimCLR was used for training the encoder in Stage-1.

\begin{table}[htbp]
	\centering
	\caption{Performance comparison across training strategies.}
	\label{tab:metrics}
	\renewcommand{\arraystretch}{1.25} 
	\begin{tabular}{lccc}
		
		\textbf{Training Strategy} &
		\shortstack{\textbf{Overall}\\\textbf{Accuracy}} &
		\shortstack{\textbf{Precision}\\\textbf{(No Motion)}} &
		\shortstack{\textbf{Recall}\\\textbf{(Severe Motion)}} \\
		\hline
		Encoder trained using SimCLR & 0.682 & 0.732 & 0.669 \\
		Fully supervised 3-class DL network         & 0.832 & 0.874 & 0.868 \\
		Proposed Method     & \textbf{0.840} & \textbf{0.951} & \textbf{0.940} \\
		\hline
	\end{tabular}
\end{table}

\subsection{Motion Grade Affinity Score (MoGrAS) analysis}
\begin{figure}[htb]
	\centering
	\caption{Violin plots of MoGrAS scores across expert ground-truth motion grades (No Motion, Subtle Motion, Severe Motion). Panels: (a) MoGrAS-NoMo, (b) MoGrAS-SuMo, (c) MoGrAS-SeMo. Red markers/line indicate per-grade medians with numeric annotations.}
	\label{fig:score_distributions}
	
	\begin{subfigure}{0.32\textwidth}
		\centering
		\includegraphics[width=\linewidth]{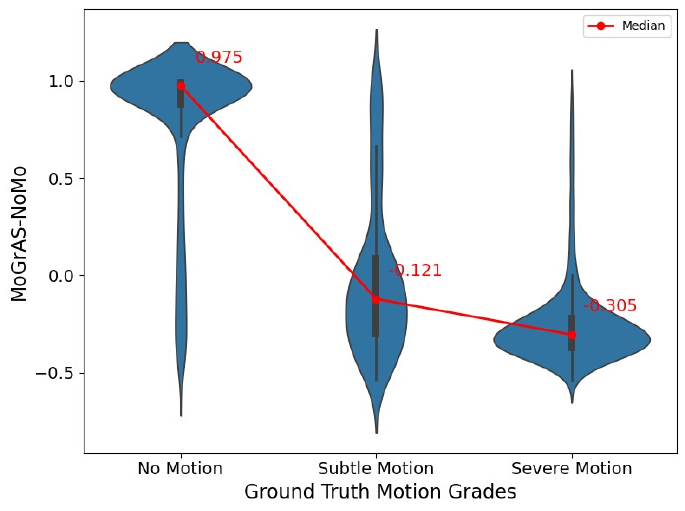}
		\caption{}%
		\label{fig:NoMo}
	\end{subfigure}
	\hfill
	\begin{subfigure}{0.32\textwidth}
		\centering
		\includegraphics[width=\linewidth]{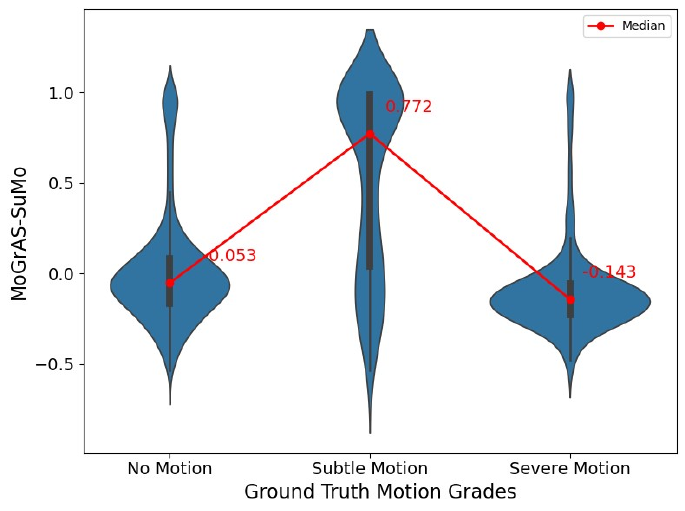}
		\caption{}%
		\label{fig:SuMo}
	\end{subfigure}
	\hfill
	\begin{subfigure}{0.32\textwidth}
		\centering
		\includegraphics[width=\linewidth]{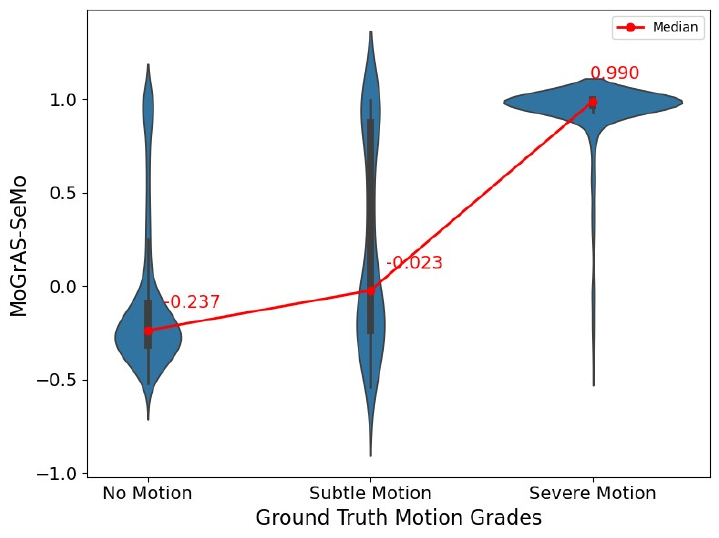}
		\caption{}%
		\label{fig:SeMo}
	\end{subfigure}
\end{figure}

MoGrAS is proposed as discussed in earlier section. An experiment is designed to ensure the relevance of the proposed motion grade affinity scoring framework. As understood from earlier discussion, each MR image has MoGrAS corresponding to each of the three motion grades: No Motion, Subtle Motion and Severe Motion. Let these be referred to as MoGrAS-NoMo, MoGrAS-SuMo and MoGrAS-SeMo respectively. In order to assess the relevance of these scores, the three scores for each MR image is plotted separately across the ground truth motion grades assigned by the expert. Refer to Figure~\ref{fig:score_distributions} for the results.

In Figure~\ref{fig:NoMo}, MoGrAS-NoMo is plotted across the ground truth motion grades for the MR images. It shows that MoGrAS-NoMo is highest for the MR images annotated as No Motion by the experts and gradually decreases as we go towards MR images with high motion. For MoGrAS-SuMo (refer Figure~\ref{fig:SuMo}), the values are highest for the MR images graded as subtle motion by the annotator (and similar for the No Motion and Severe Motion grade MR images). In case of MoGrAS-SeMo (refer Figure~\ref{fig:SeMo}), the MR images graded as severe motion has the highest values, followed by MR images graded as Subtle Motion and No Motion by the annotator.




\begin{figure}[htb]
	\centering
	\includegraphics[width=0.97\textwidth]{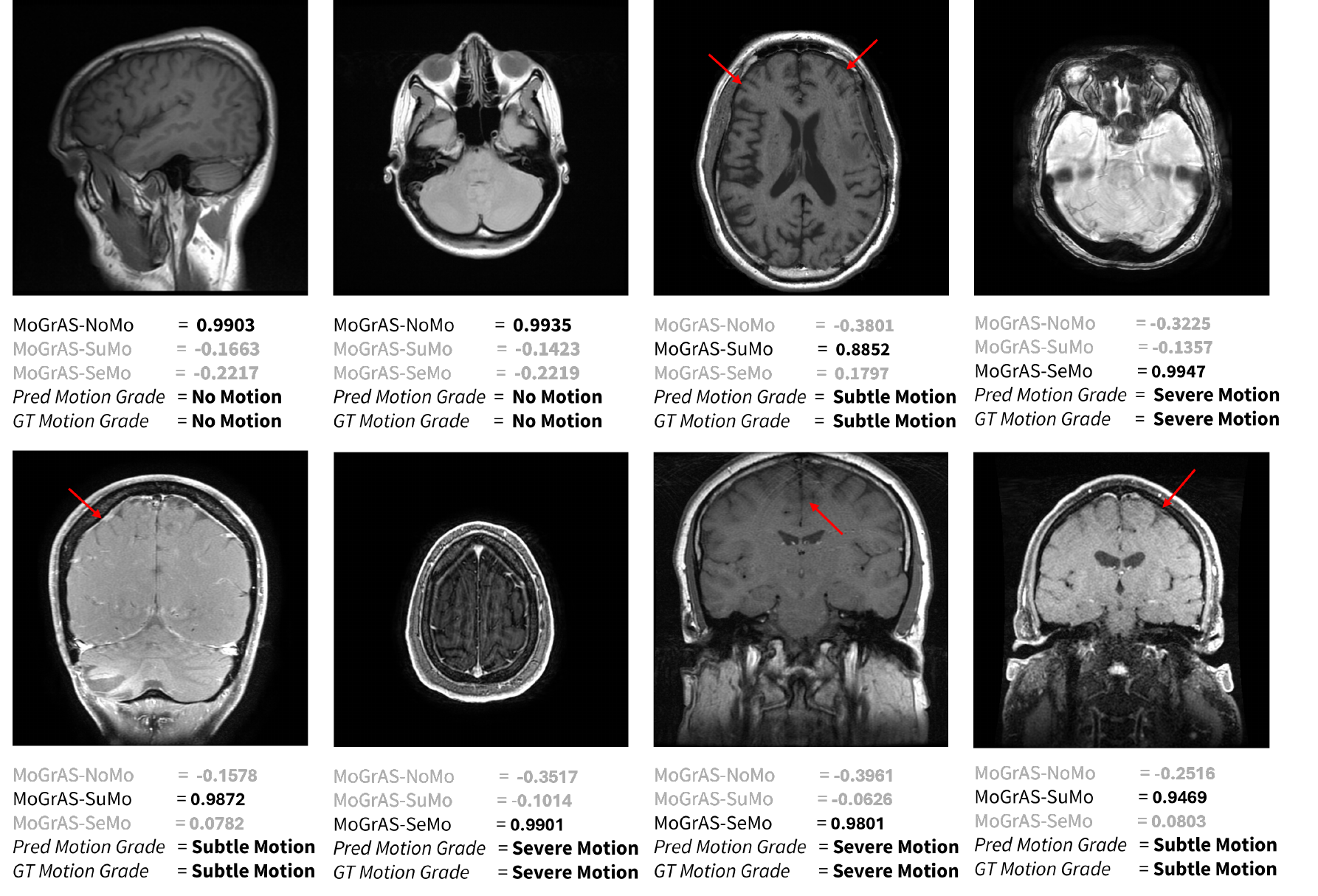}
	\caption{Examples of MR slices with predicted motion grades and associated grade affinity scores from the proposed framework. Each image displays the predicted grade, ground truth grade marked by the technologist and associated 3 grade affinity scores.}
	\label{fig:qualitative_examples}
\end{figure}

\subsection{Qualitative Examples}
Figure~\ref{fig:qualitative_examples} presents MR slices along with the corresponding predicted motion grade and the MoGrAS for each motion grade. In all examples the MoGrAS for the predicted category is higher than the MoGrAS for other motion grades, but the extent of same varies across cases. Cases with clear motion artifacts exhibit elevated MoGrAS-SeMo, while MRI scans without visible motion artifacts exhibit higher MoGrAS-NoMo. Similarly, MR images with subtle motion artifacts have been identified and scored appropriately.

\section{Discussion}
The effectiveness of the proposed method over relevant comparative methods can be understood on the basis of intra-class compactness and inter-class separation of the feature space across motion grades irrespective of variability in contrast and orientation as observed in the t-SNE plots. 
The results demonstrate that supervised contrastive learning (as compared to similar approaches as SimCLR) substantially improves feature representation quality.

MR image motion grade classification performance validates the importance of training an encoder which is able to demarcate the images from different motion grade well in feature space itself. The supervised contrastive learning based framework proposed in this work outperforms a fully supervised classifier for 3 class motion grade classification. In addition to improvement in the overall accuracy, the proposed approach achieved critical gains in class-specific metrics. In particular, recall for Severe Motion was 94\% for the proposed method compared to 86.8\% for the fully supervised 3-class classifier. Similarly, the precision for \emph{No Motion} was 95.1\% compared to 87.4\% for the fully supervised 3-class classifier, indicating that most scans predicted as artifact-free were indeed clean (reducing false alarm scenarios). Accuracy analysis indicate that the proposed method substantially reduces errors for \emph{Severe Motion} and improves calibration for \emph{Subtle Motion}, minimizing the likelihood of underestimating mild artifacts.

An important aspect of this work is the proposed motion grade affinity scoring. The goal of this metric is to explain the outcome by demonstrating conviction or tentativeness about the predicted motion grade. This derived metric was evaluated against annotator's motion grades for MR images. As discussed in last section, the proposed MoGrAS for each motion grade scales in accordance to the annotations. The nature of alignment observed, indicates that the proposed MoGrAS computation approach matches the perception of the annotator.
\section{Conclusion}
In this work, we present a framework that provides a robust and explainable solution for inline MRI quality control with respect to motion artifacts. Beyond accurately predicting the motion grade of MR images, the proposed approach introduces an affinity scoring mechanism for each grade, offering interpretability of the predicted results. This framework can be extended for artifact detection and severity assessment in medical imaging, given the availability of expert annotations.

\bibliographystyle{unsrtnat}
\bibliography{references}

@article{esteban2017mriqc,
	title={MRIQC: Advancing the automatic prediction of image quality in MRI from unseen sites},
	author={Esteban, Oscar and Birman, Daniel and Schaer, Marie and Koyejo, Oluwasanmi O and Poldrack, Russell A and Gorgolewski, Krzysztof J},
	journal={PloS one},
	volume={12},
	number={9},
	pages={e0184661},
	year={2017},
	publisher={Public Library of Science San Francisco, CA USA}
}

@article{obuchowicz2020magnetic,
	title={Magnetic resonance image quality assessment by using non-maximum suppression and entropy analysis},
	author={Obuchowicz, Rafa{\l} and Oszust, Mariusz and Bielecka, Marzena and Bielecki, Andrzej and Pi{\'o}rkowski, Adam},
	journal={Entropy},
	volume={22},
	number={2},
	pages={220},
	year={2020},
	publisher={MDPI}
}

@article{power2012spurious,
	title={Spurious but systematic correlations in functional connectivity MRI networks arise from subject motion},
	author={Power, Jonathan D and Barnes, Kelly A and Snyder, Abraham Z and Schlaggar, Bradley L and Petersen, Steven E},
	journal={Neuroimage},
	volume={59},
	number={3},
	pages={2142--2154},
	year={2012},
	publisher={Elsevier}
}

@article{reuter2015head,
	title={Head motion during MRI acquisition reduces gray matter volume and thickness estimates},
	author={Reuter, Martin and Tisdall, M Dylan and Qureshi, Abid and Buckner, Randy L and van der Kouwe, Andr{\'e} JW and Fischl, Bruce},
	journal={Neuroimage},
	volume={107},
	pages={107--115},
	year={2015},
	publisher={Elsevier}
}

@article{alexander2016subtle,
	title={Subtle in-scanner motion biases automated measurement of brain anatomy from in vivo MRI},
	author={Alexander-Bloch, Aaron and Clasen, Liv and Stockman, Michael and Ronan, Lisa and Lalonde, Francois and Giedd, Jay and Raznahan, Armin},
	journal={Human brain mapping},
	volume={37},
	number={7},
	pages={2385--2397},
	year={2016},
	publisher={Wiley Online Library}
}

@article{zaitsev2015motion,
	title={Motion artifacts in MRI: A complex problem with many partial solutions},
	author={Zaitsev, Maxim and Maclaren, Julian and Herbst, Michael},
	journal={Journal of Magnetic Resonance Imaging},
	volume={42},
	number={4},
	pages={887--901},
	year={2015},
	publisher={Wiley Online Library}
}

@article{havsteen2017movement,
	title={Are movement artifacts in magnetic resonance imaging a real problem?—a narrative review},
	author={Havsteen, Inger and Ohlhues, Anders and Madsen, Kristoffer H and Nybing, Janus Damm and Christensen, Hanne and Christensen, Anders},
	journal={Frontiers in neurology},
	volume={8},
	pages={232},
	year={2017},
	publisher={Frontiers Media SA}
}

@article{esses2018automated,
	title={Automated image quality evaluation of T2-weighted liver MRI utilizing deep learning architecture},
	author={Esses, Steven J and Lu, Xiaoguang and Zhao, Tiejun and Shanbhogue, Krishna and Dane, Bari and Bruno, Mary and Chandarana, Hersh},
	journal={Journal of Magnetic Resonance Imaging},
	volume={47},
	number={3},
	pages={723--728},
	year={2018},
	publisher={Wiley Online Library}
}

@article{vakli2023automatic,
	title={Automatic brain MRI motion artifact detection based on end-to-end deep learning is similarly effective as traditional machine learning trained on image quality metrics},
	author={Vakli, P{\'a}l and Weiss, B{\'e}la and Szalma, J{\'a}nos and Barsi, P{\'e}ter and Gyuricza, Istv{\'a}n and Kemenczky, P{\'e}ter and Somogyi, Eszter and N{\'a}rai, {\'A}d{\'a}m and G{\'a}l, Viktor and Hermann, Petra and others},
	journal={Medical Image Analysis},
	volume={88},
	pages={102850},
	year={2023},
	publisher={Elsevier}
}

@article{jimeno2024automated,
	title={Automated detection of motion artifacts in brain MR images using deep learning and explainable artificial intelligence},
	author={Jimeno, Marina Manso and Ravi, Keerthi Sravan and Fung, Maggie and Vaughan Jr, John Thomas and Geethanath, Sairam},
	journal={arXiv preprint arXiv:2402.08749},
	year={2024}
}

@article{eckerintegrating,
	title={Integrating Inline Quality Control at the MRI Scanner: Global and Local Assessment of Motion Artifacts Using Deep Learning},
	author={Ecker, Veronika and Ganz, Melanie and Eichhorn, Hannah and Marchetto, Elisa and Huelnhagen, Till and Yang, Bin and Gatidis, Sergios and K{\"u}stner, Thomas}
}

@article{khosla2020supervised,
	title={Supervised contrastive learning},
	author={Khosla, Prannay and Teterwak, Piotr and Wang, Chen and Sarna, Aaron and Tian, Yonglong and Isola, Phillip and Maschinot, Aaron and Liu, Ce and Krishnan, Dilip},
	journal={Advances in neural information processing systems},
	volume={33},
	pages={18661--18673},
	year={2020}
}

@article{maaten2008visualizing,
	title={Visualizing data using t-SNE},
	author={Maaten, Laurens van der and Hinton, Geoffrey},
	journal={Journal of machine learning research},
	volume={9},
	number={Nov},
	pages={2579--2605},
	year={2008}
}

@inproceedings{chen2020simple,
	title={A simple framework for contrastive learning of visual representations},
	author={Chen, Ting and Kornblith, Simon and Norouzi, Mohammad and Hinton, Geoffrey},
	booktitle={International conference on machine learning},
	pages={1597--1607},
	year={2020},
	organization={PmLR}
}

@inproceedings{he2016deep,
	title={Deep residual learning for image recognition},
	author={He, Kaiming and Zhang, Xiangyu and Ren, Shaoqing and Sun, Jian},
	booktitle={Proceedings of the IEEE conference on computer vision and pattern recognition},
	pages={770--778},
	year={2016}
}

\newpage
\appendix

\section{Sample Image Annotations}
\begin{figure}[htbp]
	\centering
	\includegraphics[width=0.99\textwidth]{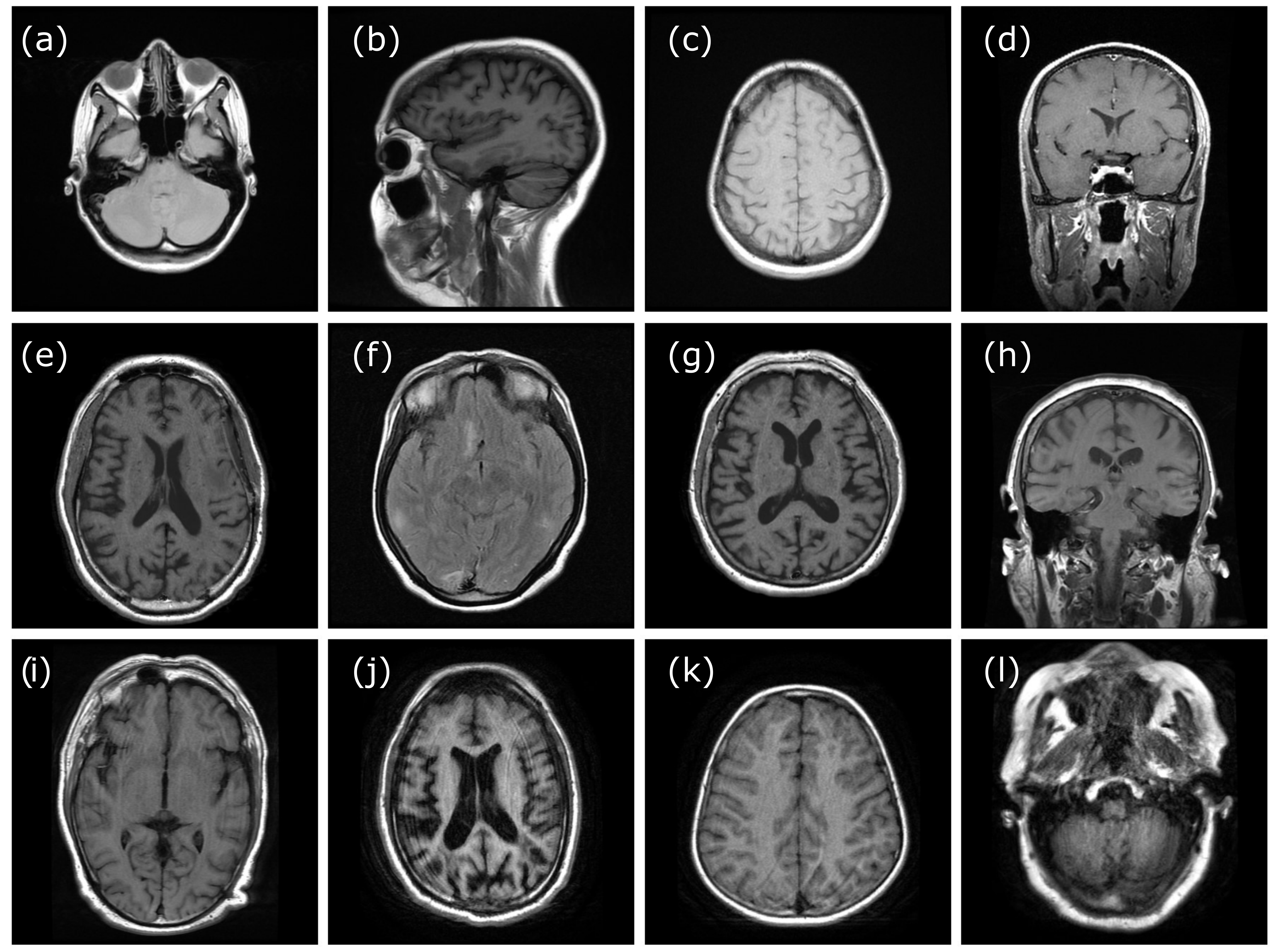}
	\caption{Representative brain MR slices illustrating the three motion grades used in this study: (a–d) No motion; (e–h) Subtle motion; (i–l) Severe motion. Examples span multiple contrasts and orientations to reflect clinical variability used in the proposed method.
	}
	\label{fig:sample_annotations}
\end{figure}

Figure~\ref{fig:sample_annotations} illustrates representative examples of the three motion grades annotated by the MR technologist.

\end{document}